\documentclass{article}

\usepackage[final]{corl_2026}
\makeatletter
\renewcommand{\@conferencelocation}{Austin, Texas, USA}
\makeatother

\usepackage{amsmath}
\usepackage{amssymb}
\usepackage{booktabs}
\usepackage{graphicx}
\usepackage{multirow}
\usepackage{url}
\usepackage{enumitem}
\usepackage{tabularx}
\usepackage{makecell}
\usepackage{array}
\usepackage{float}
\usepackage{wrapfig}
\hypersetup{hypertexnames=false}

\newcommand{\method}{\textsc{FiberTune}}
\newcommand{\source}{\textsc{Source}}
\newcommand{\calvin}{\textsc{CALVIN}}
\newcommand{\libero}{\textsc{LIBERO}}
\newcommand{\oft}{OpenVLA-OFT}
\newcommand{\pifive}{\ensuremath{\pi_{0.5}}}
\newcommand{\Rpf}{R_{\mathrm{pf}}}
\newcommand{\Ltask}{\mathcal L_{\mathrm{task}}}
\newcommand{\Lalign}{\mathcal L_{\mathrm{align}}}
\newcommand{\Lrank}{\mathcal L_{\mathrm{rank}}}
\newcommand{\Hshape}{H_{\mathrm{shape}}}

\newcolumntype{Y}{>{\raggedright\arraybackslash}X}

\title{\method: Preserving Action-Fiber Visual Residuals in Vision-Language-Action Fine-Tuning}

\author{
  \textbf{Haihao Lin$^{1,2*\dagger}$ \quad
  Xiangsheng Huang$^{3}$ \quad
  Xiao Yang$^{1}$ \quad
  Weibang Zhou$^{1}$} \\
  \textbf{Yiqi Zhang$^{4}$ \quad
  Bo Yang$^{4}$ \quad
  Simin Zeng$^{1}$ \quad
  Jiawei Yang$^{5}$} \\
  \textbf{Zhengyang Wang$^{5}$ \quad
  Jiahui Du$^{5}$} \\[0.55em]
  {\normalfont $^{1}$University of Chinese Academy of Sciences} \\
  {\normalfont $^{2}$Beijing Shengyu Qihang Robotics Technology Co., Ltd.} \\
  {\normalfont $^{3}$Zidongtaichu Foundation Model Research Center,} \\
  {\normalfont Institute of Automation, Chinese Academy of Sciences} \\
  {\normalfont $^{4}$Hebei University of Technology \quad
  $^{5}$Beijing Information Science and Technology University} \\[0.6em]
  {\normalfont\url{https://fibertune.github.io/}}
}

\hypersetup{
  pdftitle={\method: Preserving Action-Fiber Visual Residuals in Vision-Language-Action Fine-Tuning},
  pdfauthor={Haihao Lin, Xiangsheng Huang, Xiao Yang, Weibang Zhou, Yiqi Zhang, Bo Yang, Simin Zeng, Jiawei Yang, Zhengyang Wang, Jiahui Du},
  pdfsubject={CoRL 2026 camera-ready},
  pdfkeywords={Vision-Language-Action Models, Representation-Preserving Fine-Tuning, Robot Manipulation}
}

\begin{document}
\maketitle
\renewcommand*{\thefootnote}{\fnsymbol{footnote}}
\footnotetext[1]{Corresponding author: \texttt{linhaihao24@mails.ucas.ac.cn}}
\footnotetext[2]{Haihao Lin is an intern at Beijing Shengyu Qihang Robotics Technology Co., Ltd.}
\renewcommand*{\thefootnote}{\arabic{footnote}}

\begin{abstract}
Action-supervised fine-tuning of vision-language-action (VLA) policies can compress visual structure that is weakly constrained by the action objective.
We formalize this as residual visual collapse along local action fibers and propose \method{}, a training-time objective that preserves teacher-structured visual residuals without adding inference-time overhead.
\method{} uses an online action probe to filter action-predictive directions from intermediate visual-token representations. It preserves the residuals through centered directional alignment with a frozen visual teacher and effective-rank regularization.
Under identical training conditions, \method{} improves over task-loss-only fine-tuning in every one of six simulation settings spanning two benchmarks and two architectures (\pifive{} and \oft{}), as well as on physical SO-101 pick-and-place; representative gains include +10.7 pp SR(5) on long-horizon \calvin{} ABC$\rightarrow$D and physical SO-101 task success rising from 72.7\% to 78.1\%.
Representation analyses support the preservation of visual information in the residuals.
\end{abstract}


\section{Introduction}
\label{sec:introduction}

Language-conditioned robot learning has progressed from grounding language in affordances, task-conditioned imitation, and multimodal prompts to modern VLA policies that directly map observations and instructions to robot actions~\citep{ahnSayCan2022,jangBCZ2022,jiangVIMA2022,driessPalmE2023}.
Large VLA policies such as RT-1, RT-2, RoboFlamingo, OpenVLA, \oft{}, Octo, CogACT, and $\pi_0$-style policies have made it practical to adapt pretrained visual and language representations to robot control through action-supervised fine-tuning~\citep{brohanRT1,brohanRT2,liRoboFlamingo2024,kimOpenVLAOpenSourceVisionLanguageAction2024,kimFineTuningVisionLanguageActionModels2025,octo2024,liCogACT2024,blackPi0VisionLanguageActionFlow2024,physicalIntelligencePi05VisionLanguageActionModel2025}.
This adaptation is central to modern robot learning, but action-supervised losses constrain only the directions that change predicted actions --- visual factors consistent across action-equivalent states receive no direct gradient and can compress during fine-tuning.
For example, color, category, background, nearby distractors, or future-relevant object state can be visually meaningful while requiring the same short-horizon end-effector command.

We formalize this through local \emph{action fibers}: an action fiber is the set of representations that produce the same prediction under $h_\theta$; action loss constrains directions that cross fibers but leaves fiber-tangent directions --- where action-equivalent visual structure lives --- without direct gradient pressure.
Prior work has observed that VLA fine-tuning degrades inherited visual representations~\citep{kachaevDontBlindYour2025,zhangRepresentationDegradationVisionLanguageAction2025,huangMAPSPreservingVisionLanguage2025} and proposed broad preservation objectives; the action-fiber framing sharpens this to the question of which visual structure to preserve after filtering out the action-predictive component.

We introduce \method{}, a training-time residual preservation objective for VLA fine-tuning.
\method{} first estimates action-predictive feature directions using a shared online action probe.
It then applies the complementary feature-space filter to token representations, yielding a probe-filtered residual.
A frozen visual teacher supplies a training-time reference for the residual visual structure; \method{} aligns centered, normalized residual token directions to the teacher and adds an effective-rank residual prior to prevent the residual from collapsing into a small number of teacher-correlated directions.
At deployment, the policy is unchanged: the teacher, probe, and auxiliary adapter are removed.

Under the same training and evaluation conditions, \method{} improves over task-loss-only fine-tuning in all six simulation settings and on physical SO-101.
On \calvin{} ABC$\rightarrow$D, \method{} raises a \pifive{} task-loss baseline from 3.837 to 4.116 average subtasks and improves SR(5) from 61.4\% to 72.1\%.
On \libero{}, gains hold from both general and adapted checkpoints and transfer to an independent \oft{} fine-tuning stack.
On physical SO-101 pick-and-place, \method{} improves task success from 72.7\% to 78.1\%, with the largest gain on the held-out green block condition.
Representation diagnostics show increased residual teacher alignment and effective rank.

Our contributions are:
\begin{itemize}[leftmargin=1.2em,itemsep=0.15em,topsep=0.25em]
    \item We formulate residual visual collapse in VLA fine-tuning through local action fibers, clarifying why action-supervised losses can discard visually meaningful but action-equivalent factors.
    \item We propose \method{}, a training-time objective that uses action-probe filtering to apply teacher alignment only to residual visual tokens, together with an effective-rank residual prior.
    \item We demonstrate consistent gains over task-loss-only fine-tuning across \calvin{}, \libero{} (\pifive{} and \oft{}), and physical SO-101, and provide representation-level evidence of residual visual preservation.
\end{itemize}

\section{Related Work}
\label{sec:related-work}

\paragraph{Vision-language-action policies.}
Robot foundation models increasingly combine pretrained vision and language representations with action decoders for manipulation and embodied control.
Earlier systems grounded language in affordance scoring, zero-shot imitation, multimodal prompting, and embodied multimodal models~\citep{ahnSayCan2022,jangBCZ2022,jiangVIMA2022,driessPalmE2023}.
RT-1 and RT-2 demonstrated large-scale real-world robot policies and web-knowledge transfer to actions~\citep{brohanRT1,brohanRT2}.
RoboFlamingo, OpenVLA, and \oft{} adapt pretrained vision-language models into robot policies and practical fine-tuning recipes~\citep{liRoboFlamingo2024,kimOpenVLAOpenSourceVisionLanguageAction2024,kimFineTuningVisionLanguageActionModels2025}, while Octo, CogACT, and $\pi_0$-family policies explore generalist policy learning with different VLA training designs~\citep{octo2024,liCogACT2024,blackPi0VisionLanguageActionFlow2024,physicalIntelligencePi05VisionLanguageActionModel2025}.
Recent work also studies reinforced fine-tuning and long-horizon VLA execution~\citep{chenConRFT2025,huangCORFT2025,fanLongVLA2025}.
\method{} is a training-time regularizer: it acts on intermediate visual-token representations during fine-tuning and adds no parameters or computation at inference time.

\paragraph{Action-supervised policy learning.}
Strong robot policies often reduce adaptation to supervised prediction of continuous actions, action chunks, or denoising targets.
ACT and Diffusion Policy established high-performing action-sequence and diffusion-based visuomotor imitation recipes~\citep{zhaoACT2023,chiDiffusionPolicy2023}.
Recent VLA recipes extend this pattern with faster parallel decoding and action tokenization~\citep{kimFineTuningVisionLanguageActionModels2025,pertschFAST2025}.
These objectives provide direct behavioral supervision but do not specify how much non-action visual structure should remain in intermediate representations, motivating the residual-preservation question studied here.

\paragraph{Representation preservation in VLA fine-tuning.}
Several recent papers identify representation degradation as a bottleneck for VLA adaptation.
BlindVLA introduced a frozen-teacher visual alignment objective that maps VLA visual tokens through a lightweight fixed adapter and aligns them to teacher patch features with a cosine loss~\citep{kachaevDontBlindYour2025}.
MAPS preserves vision-language representations through module-wise proximity scheduling~\citep{huangMAPSPreservingVisionLanguage2025}.
Spatial Forcing aligns VLA features to spatial foundation representations~\citep{liSpatialForcingImplicit2025}.
RS-CL uses robot-state-aware contrastive regularization to shape control-relevant representations~\citep{kimContrastiveRepresentationRegularization2025}.
These methods motivate representation preservation; their objectives generally operate on full representations, teacher spaces, state metrics, or architectural constraints.
\method{} builds on this frozen-teacher alignment objective with a fixed adapter, applying teacher supervision to the probe-filtered residual and regularizing it with an effective-rank prior.

\paragraph{Information and anti-collapse principles.}
Our objective uses conditional information preservation as its design principle and implements it through trainable residual surrogates of the ideal mutual-information objective.
This places \method{} near information bottleneck views~\citep{tishby2000information,baiRethinkingLatentRedundancy2025}, LangForce's language/action conditional-information analysis of shortcut behavior~\citep{lianLangForceBayesianDecomposition2026}, and representation-rank diagnostics~\citep{garridoRankMeAssessingDownstream2022}.
The effective-rank prior draws on variance/covariance regularization in self-supervised learning~\citep{bardesVICRegVarianceInvariance2021,zbontarBarlowTwinsSelfSupervised2021} and applies this anti-collapse principle to action-filtered residual directions.
We quantify these representation-geometry properties using CKA~\citep{kornblithSimilarityNeuralNetwork2019} to measure alignment between spaces and participation-ratio effective rank to track covariance spread.

\section{Method}
\label{sec:method}

\method{} targets the part of the visual-token representation that is weakly constrained by action supervision.
We first identify this residual under a local action-fiber view, then describe the trainable probe-filtered residual used during fine-tuning.

\begin{figure}[t]
  \centering
  \includegraphics[width=\linewidth]{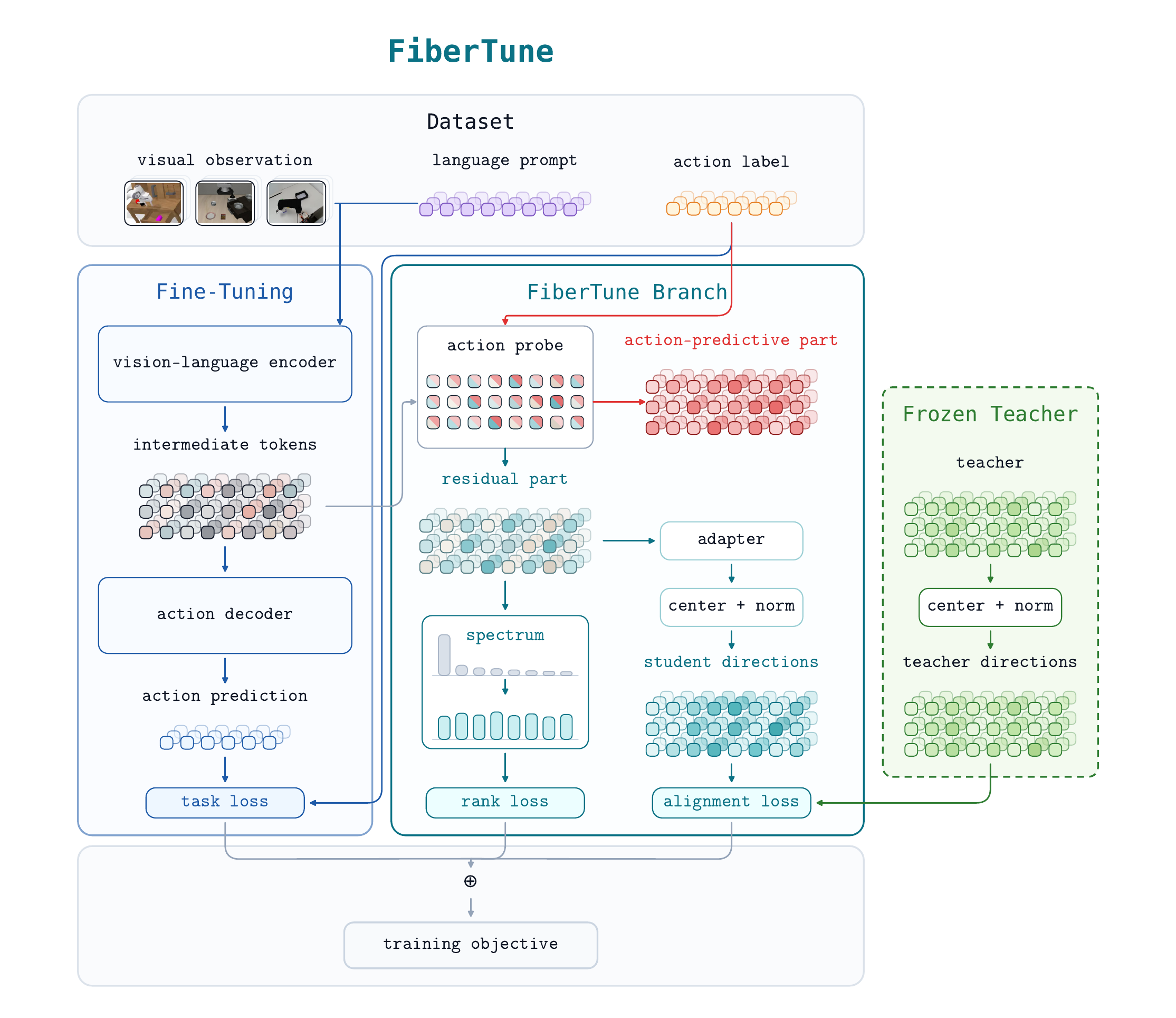}
  \caption{\textbf{\method{} overview.}
  Standard VLA fine-tuning does not supervise the action-orthogonal residual,
  allowing generalizable visual structure to collapse during adaptation.
  \method{} adds two training-time objectives---probe-filtered alignment to a frozen teacher
  and an effective-rank prior on the filtered residual---to preserve this structure;
  the teacher, probe, and auxiliary losses are removed at deployment.}
  \label{fig:method-overview}
\end{figure}

\subsection{Action fibers and residual visual collapse}
\label{sec:action-fibers}

Let $V$, $L$, and $A$ denote the visual observation, language instruction, and demonstration action.
For a sample $(v,\ell,a)$, consider an intermediate visual-token representation
$R_\theta(v,\ell)=[r_1,\ldots,r_{T_v}]\in\mathbb R^{T_v\times d}$.
A local prediction map $h_\theta$ maps these tokens to the policy output, and the task loss penalizes prediction errors through $\Ltask$.
For a fixed output value, the corresponding action fiber is the level set of representations that induce the same policy output.
Linearizing the local prediction map around the pooled visual feature $\bar R_\theta = T_v^{-1}\sum_t r_t$ gives
\begin{equation}
    h_\theta(R_\theta+\Delta R)
    \approx h_\theta(R_\theta)+J_\theta \Delta\bar r,
\end{equation}
where $\Delta\bar r = T_v^{-1}\sum_t \Delta r_t$ and $J_\theta$ is the Jacobian of the policy output with respect to $\bar R_\theta$.
Directions in $\operatorname{Row}(J_\theta)$ change the policy output to first order and are directly constrained by the task loss.
Under this pooled feature-axis linearization, directions in $\operatorname{Null}(J_\theta)$ are tangent to the action fiber to first order: they produce no first-order change in the policy output, and therefore incur no first-order task-loss change.
We denote by $U_\theta$ the feature-axis projection of $R_\theta$ onto $\operatorname{Null}(J_\theta)$.
Visual factors such as object identity, color, distractors, or future-relevant state can reside in $U_\theta$ and can be compressed during fine-tuning without a first-order penalty from the local task objective.

Let $\tau$ be a frozen visual teacher, let $Z=\tau(V)$ be its visual-token representation, and let $Z_c$ denote its token-centered form.
The teacher is introduced as a training-time reference that makes residual visual preservation optimizable from the action dataset.
\method{} uses the following conditional information objective as a preservation principle:
\begin{equation}
    \max_\theta I(U_\theta;Z_c\mid A,L)
    \quad
    \text{s.t.}
    \quad
    \Ltask(\theta)\le \epsilon,
    \label{eq:ideal}
\end{equation}
where $\epsilon$ is the allowable task-loss level.
Expanding by the chain rule gives
\begin{equation}
    I(U_\theta;Z_c\mid A,L)
    =
    H(U_\theta\mid A,L)
    -
    H(U_\theta\mid Z_c,A,L).
    \label{eq:mi-decomp}
\end{equation}
This decomposition identifies two complementary pressures.
The first term, $H(U_\theta\mid A,L)$, measures residual variability conditioned on the action and instruction.
Maximizing it encourages the residual to retain rich variation beyond what these conditions specify.
The second term, $H(U_\theta\mid Z_c,A,L)$, measures the residual uncertainty unexplained by the teacher under the same conditions.
Minimizing it encourages the residual to be predictable from the teacher's visual representation.

\subsection{From ideal residuals to a computable probe-filtered residual}
\label{sec:probe-filter}

Computing $U_\theta$ exactly requires the local Jacobian of a large nonlinear VLA policy throughout fine-tuning, which is impractical.
\method{} instead uses a scalable probe approximation to the same first-order geometry.
The approximation estimates which feature directions are predictive of demonstration actions and removes those directions from the token representation.
Concretely, \method{} fits a lightweight online linear action probe, shared across samples, from stop-gradient mean-pooled visual-token features to demonstration actions.
The probe row space estimates the action-predictive feature directions.
Writing the resulting feature filter as $P_{\mathrm{probe}}$, \method{} applies the complementary filter to the same visual-token set:
\begin{equation}
    \Rpf = R_\theta(I-P_{\mathrm{probe}}).
    \label{eq:rpf}
\end{equation}
Here $\Rpf$ is a scalable approximation to $U_\theta$, with $R_\theta P_{\mathrm{probe}}$ being the action-predictive component removed by the filter.
During the policy update, $P_{\mathrm{probe}}$ is treated as a fixed filter: auxiliary gradients flow through $\Rpf$ to $R_\theta$, while the probe is updated only via its own action-regression objective.
Because the probe is trained on pooled visual-token features, $P_{\mathrm{probe}}$ acts as shared feature-channel directions across all visual-token positions, consistent with the channel-wise structure of transformer hidden states.

\subsection{Teacher-conditioned residual alignment}
\label{sec:residual-alignment}

With $\Rpf$ in hand, we address the second pressure from Eq.~(\ref{eq:mi-decomp}): direct action supervision provides no target for action-preserving visual variation, so \method{} uses a frozen visual teacher as a training-time reference for residual visual structure.
Following BlindVLA~\citep{kachaevDontBlindYour2025}, we use a frozen teacher and a fixed randomly initialized per-token MLP adapter $g$.
\method{} applies the alignment loss to the probe-filtered residual $\Rpf$.
Before normalization, we center both representations along the token axis via $C_T=I_{T_v}-T_v^{-1}\mathbf 1\mathbf 1^\top$, removing their token means so that cosine similarity measures token-relative structure.
The loss is computed on matched image-token spans: index $t$ ranges over paired student and teacher visual tokens from the corresponding image view; language and action tokens are excluded.
Let $Y_\theta=\operatorname{norm}(C_T g(\Rpf))$ and $\tilde Z=\operatorname{norm}(Z_c)$; the alignment loss is
\begin{equation}
    \Lalign =
    - \mathbb E_{t}\left[\cos(Y_{\theta,t},\tilde Z_t)\right].
    \label{eq:align}
\end{equation}

\subsection{Effective-rank residual prior}
\label{sec:effective-rank}

The first pressure from Eq.~(\ref{eq:mi-decomp}) motivates spreading residual energy across feature directions.
Alignment alone does not guarantee this; it can satisfy the teacher objective by concentrating teacher-correlated information onto a few dominant directions.
Following covariance-spread anti-collapse objectives and rank diagnostics for learned representations~\citep{bardesVICRegVarianceInvariance2021,zbontarBarlowTwinsSelfSupervised2021,garridoRankMeAssessingDownstream2022}, we therefore regularize the shape of the batch-token residual covariance.
For the covariance spectrum $\{\lambda_i\}_{i=1}^K$ of the probe-filtered residual samples in the VLA feature space before applying $g$, where $K$ is the spectrum length used for normalization, let $p_i=\lambda_i/\sum_j\lambda_j$ and $\Hshape(\Rpf)=-(\sum_i p_i\log p_i)/\log K$.
We set $\Lrank=-\Hshape(\Rpf)$, encouraging residual energy to occupy multiple directions.
The final training objective is
\begin{equation}
    \mathcal L =
    \Ltask + \lambda_{\mathrm{align}}\Lalign + \lambda_{\mathrm{rank}}\Lrank.
    \label{eq:loss}
\end{equation}

\section{Experiments}
\label{sec:experiments}

\subsection{Experimental setup}
\label{sec:setup}

A task configuration is defined by its benchmark/task, policy, and initialization.
Within each configuration, Baseline and \method{} share the training data, number of training steps, evaluation procedure, and model-selection rule.

\paragraph{Benchmarks and metrics.}
We evaluate on two simulation benchmarks.
On \calvin{} ABC$\rightarrow$D, we evaluate multi-step language-conditioned manipulation over 1,000 trials, reporting the average number of completed subtasks and success rates at sequence lengths 1--5~\citep{mees2022calvin}.
On \libero{}, we evaluate four manipulation suites with 10 tasks per suite and 50 episodes per task, for 2,000 episodes in total~\citep{liu2023libero}.

\paragraph{Policies and initializations.}
We evaluate $\pi_{0.5}$ policies on \calvin{} and \libero{} using OpenPI-family source policies~\citep{physicalIntelligencePi05VisionLanguageActionModel2025}; fine-tuning and \libero{} evaluation use the LeRobot stack~\citep{cadenelerobot}, while \calvin{} evaluation uses the RLinf chain-evaluation bridge~\citep{yu2025rlinf,zang2025rlinf}.
We evaluate \oft{} policies on \libero{} using the original LIBERO environment~\citep{kimFineTuningVisionLanguageActionModels2025}.
For both backbones, general denotes a pretrained starting checkpoint and adapted denotes one already fine-tuned on the target benchmark; these settings test first adaptation and continued specialization, respectively.

\paragraph{Baseline comparisons.}
The task-loss baseline and \method{} differ only in their fine-tuning objective.
\method{}-specific choices --- residual layer, teacher-token span, and auxiliary loss scales --- are set per backbone before any result is collected and held constant across all rows for that backbone.
Implementation and training details are provided in Appendix~\ref{app:protocol-details}.
\subsection{Benchmark fine-tuning results}
\label{sec:main-results}

Table~\ref{tab:main} reports the main task-performance results, organized by task configuration; further details are in the appendices.

\begin{table}[t]
\centering
\caption{\textbf{Main fine-tuning results.} \method{} outperforms the task-loss baseline in all six settings. Task-performance values are from the primary single-seed runs (per-family default seed); the appendix reports multi-seed averages and Wilson 95\% confidence intervals.}
\label{tab:main}
{\scriptsize
\setlength{\tabcolsep}{4pt}
\begin{tabular}{@{}l l l l r r r@{}}
\toprule
Benchmark / task & Policy & Initialization & Metric & Baseline & \method{} & Gain \\
\midrule
\calvin{} ABC$\rightarrow$D & \pifive{} & general checkpoint & Avg.\ seq.\ len. & 0.796 & \textbf{1.012} & +0.216 \\
\calvin{} ABC$\rightarrow$D & \pifive{} & general checkpoint & SR (5) (\%) & 0.4 & \textbf{1.3} & +0.9 \\
\calvin{} ABC$\rightarrow$D & \pifive{} & adapted checkpoint & Avg.\ seq.\ len. & 3.837 & \textbf{4.116} & +0.279 \\
\calvin{} ABC$\rightarrow$D & \pifive{} & adapted checkpoint & SR (5) (\%) & 61.4 & \textbf{72.1} & +10.7 \\
\midrule
\libero{} & \pifive{} & general checkpoint & SR (\%) & 92.35 & \textbf{93.35} & +1.00 \\
\libero{} & \pifive{} & adapted checkpoint & SR (\%) & 95.75 & \textbf{97.10} & +1.35 \\
\midrule
\libero{} & \oft{} & general checkpoint & SR (\%) & 42.50 & \textbf{48.95} & +6.45 \\
\libero{} & \oft{} & adapted checkpoint & SR (\%) & 95.35 & \textbf{96.15} & +0.80 \\
\bottomrule
\end{tabular}
}
\end{table}

\begin{figure}[t]
  \centering
  \includegraphics[width=0.85\linewidth]{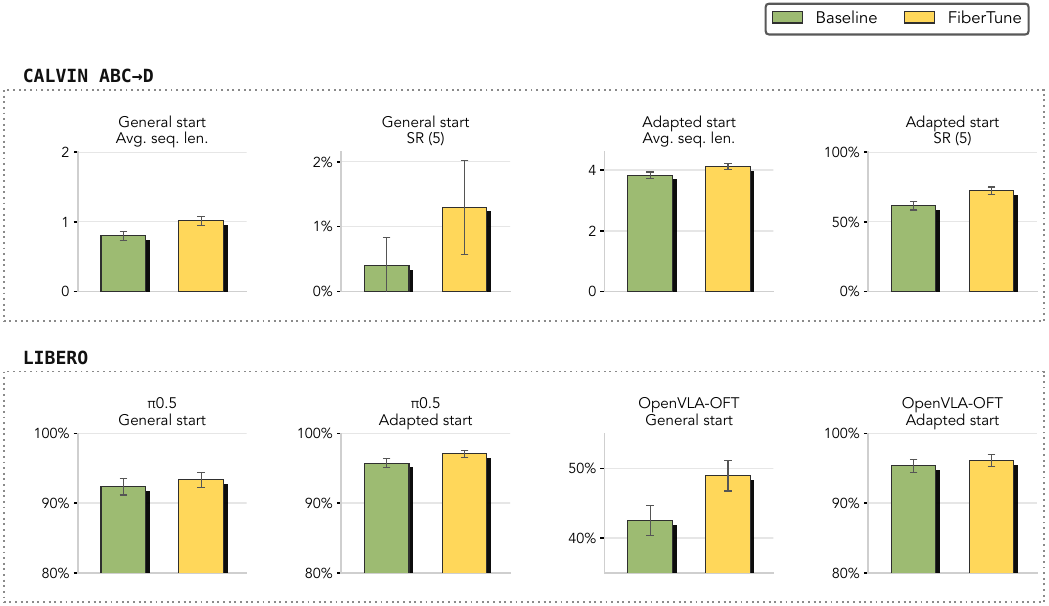}
  \caption{\textbf{Task-performance improvements with \method{}.}
  \method{} consistently outperforms the task-loss baseline across all benchmarks and policies.}
  \label{fig:main-results}
\end{figure}

\paragraph{\calvin{} long-horizon chain evaluation.}
\method{} improves average sequence length from 3.837 to 4.116 and SR(5) from 61.4\% to 72.1\%.
From a general (pre-\calvin{}) \pifive{} checkpoint, both methods remain far from task completion after 6,000 training steps (SR(5) 0.4\% vs 1.3\%).

\paragraph{\libero{} standard benchmark.}
Starting from general and adapted checkpoints, \method{} improves \pifive{} success from 92.35\% to 93.35\% and from 95.75\% to 97.10\%, respectively.

\paragraph{Results with \oft{}.}
With \oft{}, \method{} improves \libero{} success from 42.50\% to 48.95\% when initialized from a general OpenVLA checkpoint, and from 95.35\% to 96.15\% when initialized from an adapted policy.

\subsection{Real-world SO-101 pick-and-place}
\label{sec:realworld-so101}

\begin{table}[!htbp]
\centering
\caption{\textbf{SO-101 physical pick-and-place results.} \method{} improves success rates across all stages and block colors, including the held-out OOD green block.}
\label{tab:so101-realworld}
\small
\setlength{\tabcolsep}{3pt}
\begin{tabular}{@{}l c c c@{}}
\toprule
Stage / Condition & Baseline & \method{} & Gain \\
\midrule
\multicolumn{4}{@{}l}{\scriptsize\itshape Physical stage breakdown} \\
Contact & 120/128 (93.8\%) & \textbf{124/128 (96.9\%)} & +3.1 \\
Lift & 104/128 (81.2\%) & \textbf{106/128 (82.8\%)} & +1.6 \\
Task success & 93/128 (72.7\%) & \textbf{100/128 (78.1\%)} & +5.5 \\
\midrule
\multicolumn{4}{@{}l}{\scriptsize\itshape Task success by color} \\
ID colors & 78/96 (81.2\%) & \textbf{80/96 (83.3\%)} & +2.1 \\
OOD green & 15/32 (46.9\%) & \textbf{20/32 (62.5\%)} & +15.6 \\
\bottomrule
\end{tabular}
\end{table}

On the physical SO-101 pick-and-place task, \method{} improves task success from 72.7\% to 78.1\% over 128 trials per policy spanning three in-distribution block colors and one held-out green condition.
ID-color success improves from 81.2\% to 83.3\%; the held-out green condition improves from 46.9\% to 62.5\%.
On the held-out green condition the gain is largest in magnitude (46.9\%$\rightarrow$62.5\%); given the small per-condition sample ($n=32$), we treat this as suggestive physical-robot evidence for improved color generalization, with the inference-time policy unchanged.

\begin{figure}[H]
  \centering
  \includegraphics[width=0.9\linewidth]{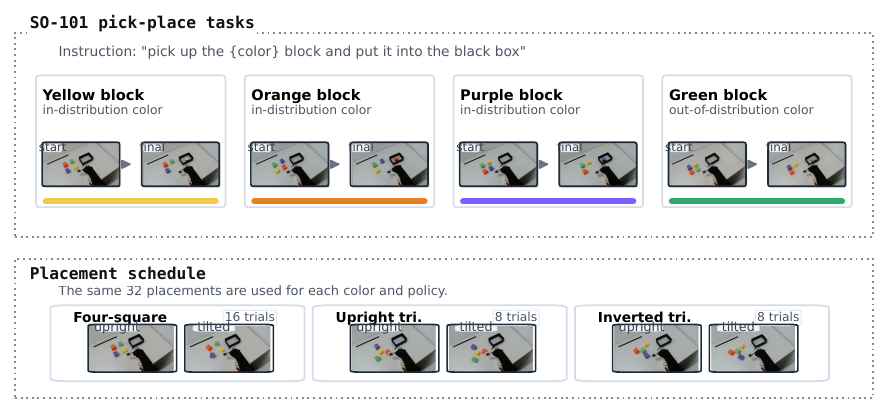}
  \caption{\textbf{Real-world SO-101 evaluation protocol.}
  The green block is the held-out OOD color; all other conditions are in-distribution.}
  \label{fig:realworld-protocol}
\end{figure}

\subsection{Representation diagnostics}
\label{sec:mechanism-diagnostics}

We examine changes in the probe-filtered residual using linear CKA~\citep{kornblithSimilarityNeuralNetwork2019} for teacher similarity and participation-ratio effective rank for covariance spread.
Table~\ref{tab:diagnostics} reports these measurements for the same task configurations as the task-performance comparisons.
Across all settings, \method{} increases both residual teacher CKA and residual effective rank.
These measurements complement the training objectives. We compute effective rank as a participation ratio and measure CKA on the probe-filtered residual before the adapter $g$; the training objectives regularize spectral entropy and optimize cosine alignment after the adapter.
The consistent increases provide representation-level evidence that \method{} preserves higher-rank, teacher-correlated residual structure, in line with the action-fiber motivation.
Further analyses are provided in Appendices~\ref{app:probe-validation} and~\ref{app:residual-diagnostics}.

\begin{table}[H]
\centering
\caption{\textbf{Representation diagnostics across task configurations.} \method{} increases both metrics in every setting. Task performance is average sequence length on \calvin{} and success rate (\%) on \libero{}.}
\label{tab:diagnostics}
\scriptsize
\setlength{\tabcolsep}{4pt}
\begin{tabular}{@{}l l c c c@{}}
\toprule
Evaluation setting & Method & Task performance & Residual CKA & Residual eff.\ rank \\
\midrule
\calvin{} / \pifive{} / general & Baseline & 0.796 & 0.112 & 4.24 \\
\calvin{} / \pifive{} / general & \method{} & \textbf{1.012} & \textbf{0.481} & \textbf{76.61} \\
\calvin{} / \pifive{} / adapted & Baseline & 3.837 & 0.202 & 8.35 \\
\calvin{} / \pifive{} / adapted & \method{} & \textbf{4.116} & \textbf{0.467} & \textbf{58.84} \\
\midrule
\libero{} / \pifive{} / general & Baseline & 92.35 & 0.154 & 6.94 \\
\libero{} / \pifive{} / general & \method{} & \textbf{93.35} & \textbf{0.680} & \textbf{78.30} \\
\libero{} / \pifive{} / adapted & Baseline & 95.75 & 0.194 & 6.37 \\
\libero{} / \pifive{} / adapted & \method{} & \textbf{97.10} & \textbf{0.414} & \textbf{44.88} \\
\midrule
\libero{} / \oft{} / general & Baseline & 42.50 & 0.380 & 39.89 \\
\libero{} / \oft{} / general & \method{} & \textbf{48.95} & \textbf{0.717} & \textbf{52.26} \\
\libero{} / \oft{} / adapted & Baseline & 95.35 & 0.349 & 35.42 \\
\libero{} / \oft{} / adapted & \method{} & \textbf{96.15} & \textbf{0.670} & \textbf{48.79} \\
\bottomrule
\end{tabular}
\end{table}

\begin{figure}[H]
  \centering
  \includegraphics[width=\linewidth]{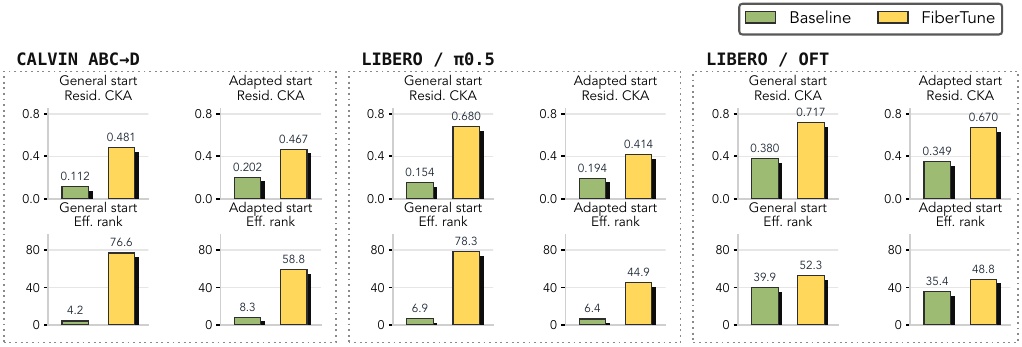}
  \caption{\textbf{Residual geometry across six task configurations.}
  \method{} increases residual CKA and effective rank in all six settings.}
  \label{fig:mechanism-diagnostics}
\end{figure}

\subsection{Component ablations}
\label{sec:component-ablations}

\begin{table}[H]
\centering
\caption{\textbf{Component ablations on \calvin{} and \libero{} using adapted \pifive{} checkpoints.} The residual rank-only variant retains the learned action probe and filter.}
\label{tab:ablation}
\scriptsize
\begin{tabular}{lllrrr}
\toprule
Variant & Teacher signal & Rank prior & \makecell{CALVIN average\\sequence length} & \makecell{CALVIN\\SR(5) (\%)} & \makecell{LIBERO\\success rate (\%)} \\
\midrule
Baseline & none & no & 3.837 & 61.4 & 95.75 \\
Full-token alignment & full tokens & no & 3.685 & 58.3 & 96.55 \\
\makecell[l]{Full-token alignment\\+ rank regularization} & full tokens & yes & 3.956 & 65.6 & 95.15 \\
Residual alignment & probe-filtered residual & no & 3.980 & 66.2 & 96.25 \\
Residual rank only & none & yes & 4.064 & 70.5 & 95.80 \\
\method{} & probe-filtered residual & yes & \textbf{4.116} & \textbf{72.1} & \textbf{97.10} \\
\bottomrule
\end{tabular}
\end{table}

Among the probe-filtered single-loss variants, rank regularization recovers most of the gain on \calvin{}, whereas teacher alignment performs better on \libero{}.
The full method achieves the best results on both benchmarks, outperforming both filtered single-loss variants and full-token alignment with rank regularization.
These results support targeting visual preservation to the probe-filtered residual during action-supervised fine-tuning.
Additional \calvin{} filtering controls are reported in Appendix~\ref{app:component-details}.

\section{Conclusion}
\label{sec:conclusion}

Action-supervised fine-tuning constrains directions that cross action fibers but leaves fiber-tangent visual structure --- where action-equivalent factors such as object color, category, and future-relevant state reside --- without direct gradient pressure.
\method{} addresses this by filtering action-predictive directions before applying teacher alignment and by maintaining the effective rank of the residual, both at training time only.
Across \calvin{}, \libero{}, and physical SO-101, \method{} consistently outperforms task-loss-only fine-tuning; representation diagnostics and component ablations support preserving residual visual structure through alignment and rank regularization.
Explicitly regularizing the action-orthogonal residual subspace is a lightweight complement to standard task-loss fine-tuning with meaningful impact on behavioral generalization.

\section*{Limitations}
\label{sec:limitations}

\method{} introduces some training-time overhead: a frozen teacher, an online probe, and auxiliary losses are required during fine-tuning, although inference remains unchanged.
The probe-filtered residual is a first-order, pooled-feature approximation to the local action-fiber residual; in settings where action prediction is highly nonlinear or where visual token positions contribute heterogeneously to the action target, the approximation may break down.
We use benchmark-specific fixed auxiliary-to-task loss scales selected from short calibration runs; these scales are not automatically transferred across benchmarks, and broader automatic balancing is left for future work.
Finally, physical-robot evidence is limited to a single task and manipulator; failure modes under more diverse real-world conditions remain uncharacterized and will require future investigation.

\acknowledgments{This work was supported in part by the Regional Innovation and Development Joint Fund of the National Natural Science Foundation of China under Grant U24A6005.}

\clearpage
\bibliography{references}

\clearpage
\appendix
\setcounter{figure}{0}
\setcounter{table}{0}
\renewcommand{\thefigure}{\Alph{section}.\arabic{figure}}
\renewcommand{\thetable}{\Alph{section}.\arabic{table}}
\renewcommand{\theHfigure}{appendix.\Alph{section}.\arabic{figure}}
\renewcommand{\theHtable}{appendix.\Alph{section}.\arabic{table}}

\section{Technical details}
\label{app:technical-details}

\subsection{From ideal orthogonal residuals to probe-filtered residuals}
\label{app:ideal-implemented-residuals}

Let $A$ denote the demonstration action. For a sample $(v,\ell,a)$, \method{} operates on an intermediate visual-token representation
\begin{equation}
    R_\theta(v,\ell)=[r_1,\ldots,r_{T_v}]\in\mathbb R^{T_v\times d},
\end{equation}
and the local prediction map $h_\theta$ is trained with
\begin{equation}
    \Ltask(\theta)
    =
    \mathbb E\!\left[
    c_{\mathrm{task}}(h_\theta(R_\theta),S)
    \right].
\end{equation}
Here $S$ is the supervised policy-output target.
For a fixed output value $s$, the action fiber is the level set
\begin{equation}
    \mathcal F_{s,\theta}=\{R:h_\theta(R)=s\}.
\end{equation}
Let $Z=\tau(V)$ be the visual-token representation of the frozen teacher and let $Z_c=C_TZ$ be its token-centered form.
The ideal preservation objective in the main paper uses the standard information-theoretic decomposition~\citep{coverElementsInformationTheory2006}
\begin{equation}
I(U_\theta;Z_c\mid A,L)
=H(U_\theta\mid A,L)-H(U_\theta\mid Z_c,A,L),
\end{equation}
which separates residual spread from teacher predictability.

We formalize the action-fiber residual with a first-order approximation of the supervised policy-output map.
Let $\bar R_\theta=T_v^{-1}\sum_t R_{\theta,t}$ be the pooled visual-token representation and let $J_\theta$ denote the Jacobian of the policy output with respect to the pooled feature $\bar R_\theta$.
Under the first-order geometry used in the main text, the ideal residual projector is
\begin{equation}
    Q_\theta
    =
    I - J_\theta^\top (J_\theta J_\theta^\top)^\dagger J_\theta,
    \qquad
    U_\theta = R_\theta Q_\theta .
\end{equation}
The projector $Q_\theta$ keeps the component of $R_\theta$ in the first-order tangent space of the local action fiber.
During training, \method{} uses a probe-defined feature filter in place of this Jacobian projector.
Let $A_{\mathrm{probe}}$ denote the demonstration-action vector used as the probe target, taken from the first time step in our experiments.
The shared online probe is trained on detached pooled visual-token features using the regression objective
\begin{equation}
    \mathcal L_{\mathrm{probe}}(W)
    =
    \mathbb E\|A_{\mathrm{probe}}-W\bar R_\theta\|^2.
\end{equation}
Its current weights $W$ define the ridge-stabilized feature filter
\begin{equation}
    P_{\mathrm{probe}}
    =
    W^{\top}(WW^{\top}+\eta I)^{-1}W.
\end{equation}
\method{} applies the complementary feature filter to the same visual-token set:
\begin{equation}
    \Rpf = R_\theta(I-P_{\mathrm{probe}}).
\end{equation}
The alignment loss and effective-rank prior both act on $\Rpf$ during training.

\subsection{Directional alignment objective}
\label{app:directional-alignment}

The directional alignment term acts on the probe-filtered residual $\Rpf$.
The adapter $g$ maps each residual token to teacher feature space; the centering matrix $C_T=I_{T_v}-T_v^{-1}\mathbf 1\mathbf 1^\top$ removes the token mean before row-wise normalization.
The cosine term therefore compares token-relative residual structure.
The resulting directions and alignment loss are
\begin{equation}
  \begin{aligned}
    Y_\theta &= \operatorname{norm}\!\left(C_T g(\Rpf)\right)\\
    \tilde Z &= \operatorname{norm}(Z_c)=\operatorname{norm}\!\left(C_TZ\right)\\
    \Lalign &=
    - \mathbb E_{t}\left[\cos(Y_{\theta,t},\tilde Z_t)\right].
  \end{aligned}
\end{equation}
A vMF-style directional likelihood over normalized residual directions gives the cosine form used in this objective~\citep{mardiaDirectionalStatistics1999}.
For a fixed concentration $\kappa>0$,
\begin{equation}
    \log p(Y_{\theta,t}\mid \tilde Z_t,\kappa)
    =
    \kappa\,Y_{\theta,t}^{\top}\tilde Z_t
    + \mathrm{const}(\kappa),
\end{equation}
so maximizing the directional likelihood is equivalent to minimizing $-\cos(Y_{\theta,t},\tilde Z_t)$.
The alignment term is therefore a directional likelihood objective on the action-filtered residual.

\subsection{Effective-rank prior}
\label{app:effective-rank-prior}

For the probe-filtered residual $\Rpf$, the effective-rank prior uses the empirical covariance over batch-token residual samples in the VLA feature space before applying $g$.
For a mini-batch $\mathcal B$ with $N=|\mathcal B|T_v$ batch-token samples, SVD of the centered sample matrix returns a spectrum of length $K=\min(N,d)$.
With trace-normalized eigenvalues $p_i$, the normalized score $-(\sum_i p_i\log p_i)/\log K$ is scale-free and penalizes concentration into a small number of residual directions.
Concretely, for covariance eigenvalues $\{\lambda_i\}$,
\begin{equation}
  \begin{aligned}
    p_i &= \frac{\lambda_i}{\sum_j \lambda_j}\\
    \Hshape(\Rpf) &= -\frac{1}{\log K}\sum_i p_i\log p_i\\
    \Lrank &= -\Hshape(\Rpf).
  \end{aligned}
\end{equation}
The rank term therefore favors residual covariance spectra that remain spread across multiple directions.
For numerical stability, the implementation floors covariance eigenvalues before trace normalization.

\subsection{Probe row-space consistency}
\label{app:probe-row-space-consistency}

Under a centered local linear model, the action probe estimates the row space of the map predicting its demonstration-action target.
\paragraph{Proposition.}
Suppose the centered pooled representation and action target satisfy
\begin{equation}
    \tilde R=\bar R_\theta-\mathbb E[\bar R_\theta],
    \qquad
    \tilde A_{\mathrm{probe}}=A_{\mathrm{probe}}-\mathbb E[A_{\mathrm{probe}}],
\end{equation}
and
\begin{equation}
    \tilde A_{\mathrm{probe}}=J_0\tilde R+\varepsilon,
    \qquad
    \mathbb E[\varepsilon\mid \tilde R]=0,
\end{equation}
with $\mathbb E[\tilde R\tilde R^\top]$ full rank.
Then the unregularized population least-squares probe
\begin{equation}
    W_0^*
    =
    \arg\min_W \mathbb E\|\tilde A_{\mathrm{probe}}-W\tilde R\|^2
\end{equation}
recovers the local action-predictive row space.

\paragraph{Proof.}
The population solution is
\begin{equation}
\begin{aligned}
W_0^*
&=
\mathbb E[\tilde A_{\mathrm{probe}}\tilde R^\top]\,
\mathbb E[\tilde R\tilde R^\top]^{-1} \\
&=
J_0\mathbb E[\tilde R\tilde R^\top]\,
\mathbb E[\tilde R\tilde R^\top]^{-1}
=J_0 .
\end{aligned}
\end{equation}
Thus the probe row space equals the row space of this target map under the stated local model.
With finite samples, ridge stabilization gives the filter $P_{\mathrm{probe}}$ used by \method{}.

\paragraph{Empirical limit.}
Let $\widehat W_n$ be the ridge probe fitted from $n$ detached pooled visual-token samples under the same local linear model.
If the empirical cross-covariance and feature covariance converge to their population values, the feature covariance remains nonsingular, and the probe-fitting ridge coefficient $\lambda_n\to 0$, then $\widehat W_n\to W_0^*$.
For a fixed positive filter ridge coefficient $\eta$, the corresponding filter converges to the ridge-stabilized filter defined by the population probe $W_0^*$.

\section{Protocol details}
\label{app:protocol-details}

\subsection{Implementation stacks and seed conventions}
\label{app:implementation-stacks}

Our experiments span two VLA model families whose public training and evaluation toolchains differ, so we record the implementation stack used in each setting.
The \pifive{} family was released through Physical Intelligence/OpenPI~\citep{blackPi0VisionLanguageActionFlow2024,physicalIntelligencePi05VisionLanguageActionModel2025}.
We fine-tune these policies with Hugging Face LeRobot~\citep{cadenelerobot}, using the same implementation for the task-loss baseline and \method{}.
For \libero{}, we use LeRobot evaluation code; for SO-101, we use its real-robot recording pipeline.
For \calvin{}, policy checkpoints are first converted to the OpenPI format and evaluated with RLinf/OpenPI chain-evaluation code~\citep{yu2025rlinf,zang2025rlinf}.
The \oft{} experiments use the OpenVLA-OFT codebase, including its data pipeline and \libero{} evaluation code~\citep{kimFineTuningVisionLanguageActionModels2025}.

The seeds used in our experiments are the public defaults of the corresponding implementation toolchains.
OpenVLA-OFT uses seed 7 as the default \libero{} evaluation seed in its released evaluation script.\footnote{\url{https://raw.githubusercontent.com/moojink/openvla-oft/main/experiments/robot/libero/run_libero_eval.py}}
OpenPI uses seed 42 as the default training seed in its public training configuration,\footnote{\url{https://raw.githubusercontent.com/Physical-Intelligence/openpi/main/src/openpi/training/config.py}}
and LeRobot uses seed 1000 as the default seed in its released training configuration\footnote{\url{https://raw.githubusercontent.com/huggingface/lerobot/main/src/lerobot/configs/train.py}}
and evaluation configuration.\footnote{\url{https://raw.githubusercontent.com/huggingface/lerobot/main/src/lerobot/configs/eval.py}}

\subsection{Training and evaluation settings}
\label{app:matched-protocol-summary}

Table~\ref{tab:protocol-details} lists the initialization, training data, and evaluation settings shared by the task-loss baseline and \method{} in each comparison.
Starting checkpoints are detailed in Table~\ref{tab:source-provenance}, and training steps and learning-rate schedules in Table~\ref{tab:hyperparameters}.

\begin{table}[!htbp]
\centering
\caption{Initialization, training data, and evaluation settings shared by the task-loss baseline and \method{} in each comparison.}
\label{tab:protocol-details}
\scriptsize
\setlength{\tabcolsep}{3pt}
\begin{tabularx}{\linewidth}{@{}l l l Y Y@{}}
\toprule
Evaluation setting & Model & Initialization & Training data & Evaluation settings \\
\midrule
\calvin{} ABC$\rightarrow$D & \pifive{} & adapted & CALVIN ABC & 1000 CALVIN chain-evaluation trials \\
\calvin{} ABC$\rightarrow$D & \pifive{} & general & CALVIN ABC & 1000 CALVIN chain-evaluation trials \\
\libero{} & \pifive{} & general & \libero{} task suites & 10 tasks per suite and 50 episodes per task \\
\libero{} & \pifive{} & adapted & \libero{} task suites & 10 tasks per suite and 50 episodes per task \\
\libero{} & \oft{} & general & \libero{} task suites & 10 tasks per suite and 50 episodes per task \\
\libero{} & \oft{} & adapted & \libero{} task suites & 10 tasks per suite and 50 episodes per task \\
SO-101 pick-and-place & \pifive{} & real-world source & 90 SO-101 demonstrations; yellow/orange/purple ID colors; green held out for OOD evaluation & 128 physical trials per policy \\
\bottomrule
\end{tabularx}
\par\vspace{2pt}
{\scriptsize CALVIN normalization and the chain-evaluation bridge use the released RLinf/OpenPI CALVIN asset.}
\end{table}

\subsection{Source-policy provenance}
\label{app:source-provenance}

Table~\ref{tab:source-provenance} records the starting-policy provenance for the main experiments.

\begin{table}[!htbp]
\centering
\caption{Starting-policy provenance for each task configuration.}
\label{tab:source-provenance}
\scriptsize
\setlength{\tabcolsep}{3pt}
\begin{tabularx}{\linewidth}{@{}l l Y@{}}
\toprule
Task configuration & Training start & Source-policy provenance \\
\midrule
\calvin{} / \pifive{} / general & general checkpoint & \texttt{lerobot/pi05\_base}. \\
\calvin{} / \pifive{} / adapted & adapted checkpoint & \texttt{RLinf/RLinf-Pi05-CALVIN-ABC-D-SFT}. \\
\libero{} / \pifive{} / general & general checkpoint & \texttt{lerobot/pi05\_base}. \\
\libero{} / \pifive{} / adapted & adapted checkpoint & \texttt{lerobot/pi05\_libero}. \\
\libero{} / \oft{} / general & general checkpoint & \texttt{openvla/openvla-7b}. \\
\libero{} / \oft{} / adapted & adapted checkpoint & \makecell[l]{\texttt{moojink/openvla-7b-oft-}\\\texttt{finetuned-libero-spatial-object-goal-10}.} \\
SO-101 / \pifive{} & real-world source policy & Trained from \texttt{lerobot/pi05\_base} on the 90-demonstration SO-101 source set. \\
\bottomrule
\end{tabularx}
\end{table}

\subsection{Implementation details for residual objectives}
\label{app:implementation-protocol}

Table~\ref{tab:implementation-protocol} records the implementation choices behind \method{}'s residual objectives.

\begin{table}[!htbp]
\centering
\caption{Implementation details for \method{} residual losses in the main experiments. ``Same'' marks an \oft{} setting identical to the \pifive{} column.}
\label{tab:implementation-protocol}
\scriptsize
\setlength{\tabcolsep}{2pt}
\begin{tabularx}{\linewidth}{@{}lYY@{}}
\toprule
Field & \pifive{} experiments & \oft{} experiments \\
\midrule
Visual-token source & Prefix-tower image-token slice from an intermediate policy layer & VLA hidden-state image-token span from an intermediate policy layer \\
Layer used in reported runs & Prefix tower layer 8 & Layer 16 \\
Teacher & Frozen RADIO visual teacher, \texttt{c-radio\_v3-l}~\citep{heinrichRADIOv25Improved2025} & Same \\
Teacher alignment & Frozen-teacher, fixed-adapter token cosine alignment applied to $\Rpf$ & Same \\
View used for residual objectives & Primary camera view & Same \\
Token pairing & Student image-token count is matched to teacher patch-token count for the selected view; mismatched tokenizations require an explicit adapter & Same \\
Policy-output target & Flow velocity & Continuous action vector \\
Probe target & First-step clean demonstration action & First-step continuous action vector \\
Probe fitting & Shared online no-bias linear probe on detached mean-pooled visual-token features; updated before filtering & Same \\
Ridge filter & $P_{\mathrm{probe}}=W^\top(WW^\top+\eta I)^{-1}W$, $\eta=10^{-4}$ & Same \\
Adapter $g$ & Randomly initialized MLP adapter mapping each token position independently into teacher feature space; adapter hidden width 2048; frozen after initialization in reported experiments & Same \\
Auxiliary gradient path & Gradients pass through fixed $P_{\mathrm{probe}}$ and fixed $g$ to the policy representation; probe and adapter $g$ are not optimized by $\Lalign$ & Same \\
Covariance samples for eff.-rank prior & Batch-token samples from $\Rpf$ before applying $g$ & Same \\
\bottomrule
\end{tabularx}
\end{table}

\subsection{Hyperparameters}
\label{app:hyperparameters}

\begin{table}[!htbp]
\centering
\caption{Hyperparameters for the main experiments.}
\label{tab:hyperparameters}
\scriptsize
\setlength{\tabcolsep}{2.5pt}
\begin{tabularx}{\linewidth}{@{}Y c c Y Y@{}}
\toprule
Evaluation setting & Batch / accum. / GPUs & Training steps & LR schedule & Aux coefficients \\
\midrule
\calvin{} / \pifive{} / general & 8 / 4 / 1 & 6,000 steps & $2.5{\times}10^{-5}\!\rightarrow\!2.5{\times}10^{-6}$ cosine; 1k warmup & 0.0125 / 0.00625; 1k aux warmup; decay 3k--6k to 0.5 \\
\calvin{} / \pifive{} / adapted & 8 / 4 / 1 & 6,000 steps & $2.5{\times}10^{-5}\!\rightarrow\!2.5{\times}10^{-6}$ cosine; 1k warmup & 0.0125 / 0.00625; 1k aux warmup; decay 3k--6k to 0.5 \\
\libero{} / \pifive{} / general & 8 / 4 / 2 & 6,000 steps & $5{\times}10^{-5}\!\rightarrow\!5{\times}10^{-6}$ cosine; 1k warmup & 0.0125 / 0.00625; 1k aux warmup; decay 3k--6k to 0.5 \\
\libero{} / \pifive{} / adapted & 8 / 4 / 1 & 6,000 steps & $2.5{\times}10^{-5}\!\rightarrow\!2.5{\times}10^{-6}$ cosine; 1k warmup & 0.0125 / 0.00625; 1k aux warmup; decay 3k--6k to 0.5 \\
\libero{} / \oft{} / general & 8 / 1 / 1 & 20,000 steps & $10^{-4}$; 1k warmup + constant & 0.0125 / 0.00625; 1k aux warmup; decay 10k--20k to 0.5 \\
\libero{} / \oft{} / adapted & 8 / 1 / 1 & 15,000 steps & $10^{-4}$; 1k warmup + constant & 0.0125 / 0.00625; 1k aux warmup; decay 10k--20k to 0.5 \\
SO-101 / \pifive{} / pick-and-place policy & 8 / 8 / 1 & 6,000 steps & $5{\times}10^{-5}\!\rightarrow\!5{\times}10^{-6}$ cosine; 1k warmup & 0.003125 / 0.0015625; 3k aux warmup; decay after 4.5k to 0.5 \\
\bottomrule
\end{tabularx}
\end{table}
Batch denotes per-device batch size, and the optimizer effective batch is batch size $\times$ accumulation $\times$ GPU count.
The residual probe and residual covariance estimators are computed on each forward micro-batch.
Reported runs used NVIDIA GPUs with at least 48GB of device memory; the GPU count for each comparison is listed in the table.

\paragraph{Auxiliary-coefficient selection.}
We choose the auxiliary coefficients from short calibration runs so that the auxiliary-to-task loss ratio,
\begin{equation}
\rho_{\mathrm{aux}}
=
\frac{
\lambda_{\mathrm{align}}|\Lalign|
+\lambda_{\mathrm{rank}}|\Lrank|
}{
\Ltask
},
\end{equation}
is approximately $3$--$6\%$ near the end of calibration.

\subsection{Training-time cost}
\label{app:training-cost}

We profile task-loss-only and \method{} training on one NVIDIA RTX 6000 Ada GPU with compilation disabled, using configurations matched to the reported training setup.
Gradient accumulation is four for \pifive{} and one for \oft{}.
Table~\ref{tab:training-cost} reports optimizer-step time and GPU memory for each pair.
\begin{table}[!htbp]
\centering
\caption{Training cost under the stated single-GPU configurations. Percentages in parentheses indicate the increase in optimizer-step time relative to task-loss-only training for the same model.}
\label{tab:training-cost}
\begin{tabular}{llrr}
\toprule
Model & Training objective & Time per optimizer step (s) & GPU memory (GiB) \\
\midrule
\pifive{} & Task-loss only & 9.671 & 36.978 \\
\pifive{} & \method{} & 10.548 (+9.07\%) & 38.363 \\
\oft{} & Task-loss only & 2.383 & 29.961 \\
\oft{} & \method{} & 2.603 (+9.24\%) & 30.753 \\
\bottomrule
\end{tabular}
\end{table}

\subsection{Representation metrics and probe fitting}
\label{app:diagnostic-protocol}

The representation diagnostics reported in the main paper use a probe refit for analysis, separate from the online probe weights saved during training.
This probe predicts the 7D per-step action vector.
We use four-fold cross-fitting for \calvin{} and \pifive{} \libero{}, and in-sample fitting for \oft{}.
Here, $\Rpf$ denotes the residual formed by the refit probe.
Residual CKA is linear CKA between $\Rpf$ and the matched teacher features before the adapter $g$.
Residual effective rank is the participation-ratio statistic
$d_{\mathrm{eff}}=(\sum_i\lambda_i)^2/\sum_i\lambda_i^2$
computed from the residual covariance spectrum.
The training objective uses the normalized spectral entropy $\Hshape$ from Appendix~\ref{app:effective-rank-prior}; the reported effective rank is a complementary summary of the same covariance spectrum.

\subsection{BlindVLA-style full-token alignment baseline}
\label{app:blindvla-style-full-token}

The main-paper ``Full-token alignment'' ablation implements BlindVLA-style frozen-teacher visual alignment in our \pifive{} / LeRobot training setup~\citep{kachaevDontBlindYour2025}.
It uses the same frozen RADIO teacher and fixed randomly initialized per-token adapter as \method{}.
The adapter maps unfiltered intermediate visual tokens into the teacher feature space, where student and teacher tokens are directly L2-normalized for cosine alignment.
Within each benchmark, this variant shares the initialization, training data, budget, and evaluator of the other component ablations.

BlindVLA studies visual alignment in OOD-oriented settings, including Simpler-based generalization evaluations and the VL-Think suite~\citep{kachaevDontBlindYour2025}.
Our ablations examine full-token alignment during continued fine-tuning on \calvin{} and \libero{}.
Relative to task-loss-only fine-tuning, it decreases \calvin{} SR(5) and improves \libero{} success (Table~\ref{tab:ablation}).
One possible explanation is that the balance between visual preservation and action adaptation varies across tasks and starting policies.
Full-token alignment can retain useful visual information while also constraining feature directions involved in action prediction.
The benefits and costs of these constraints may therefore differ across adaptation settings.

\method{} changes both the alignment target and its geometry.
An online action probe filters action-predictive feature directions from the visual tokens, directing teacher supervision toward the remaining residual.
After projection through the adapter, the residual tokens and teacher tokens are separately centered along the token axis before L2 normalization.
Uncentered cosine alignment can be dominated by shared mean directions within each token set; centering removes these means and focuses alignment on token-relative visual structure.
These changes may help preserve visual information while reducing constraints on action-predictive directions and sensitivity to mean-feature differences.
The full-token versus residual-alignment comparison captures their combined effect.
\method{} additionally applies spectral-entropy regularization to the residual covariance before the adapter to encourage variation across feature directions.
\section{Additional analyses}
\label{app:additional-analyses}

\subsection{CALVIN results across matched seeds}
\label{app:calvin-paired-seeds}

Table~\ref{tab:calvin-paired-seeds} reports \calvin{} results across matched seeds for the adapted starting checkpoint as a robustness check, using the same settings and 6,000 training steps as the main \calvin{} adapted comparison.

\begin{table}[!htbp]
\centering
\caption{\calvin{} results across matched seeds for the adapted starting checkpoint. SR1--SR5 report the percentage of rollouts that complete at least 1--5 consecutive subtasks.}
\label{tab:calvin-paired-seeds}
\scriptsize
\setlength{\tabcolsep}{2.7pt}
\begin{tabular}{@{}c l c c c c c c@{}}
\toprule
Seed & Method & Avg. & SR1 & SR2 & SR3 & SR4 & SR5 \\
\midrule
\multirow{2}{*}{5}
& Baseline & 3.682 & 90.7 & 81.1 & 73.1 & 65.6 & 57.7 \\
& \method{} & \textbf{4.044} & \textbf{93.0} & \textbf{85.9} & \textbf{80.2} & \textbf{75.2} & \textbf{70.1} \\
\midrule
\multirow{2}{*}{6}
& Baseline & 3.980 & 93.0 & 85.6 & 78.9 & 73.2 & 67.3 \\
& \method{} & \textbf{4.105} & \textbf{94.2} & \textbf{87.4} & \textbf{81.8} & \textbf{76.6} & \textbf{70.5} \\
\midrule
\multirow{2}{*}{7}
& Baseline & 3.837 & 93.2 & 84.7 & 75.9 & 68.5 & 61.4 \\
& \method{} & \textbf{4.116} & \textbf{94.3} & \textbf{87.3} & \textbf{80.9} & \textbf{77.0} & \textbf{72.1} \\
\midrule
\multirow{2}{*}{Mean}
& Baseline & 3.833 & 92.3 & 83.8 & 76.0 & 69.1 & 62.1 \\
& \method{} & \textbf{4.088} & \textbf{93.8} & \textbf{86.9} & \textbf{81.0} & \textbf{76.3} & \textbf{70.9} \\
\bottomrule
\end{tabular}
\end{table}

\subsection{\libero{} per-suite success rates}
\label{app:libero-per-suite}

Table~\ref{tab:libero-suite-breakdown} provides suite-level success rates for the \libero{} configurations with available breakdowns.
The \pifive{} / adapted block spans seeds 7, 42, and 1000, with a three-seed average gain of +0.70 percentage points.
For the 20,000-step \oft{} / general-start setting, the $+6.45$ pp aggregate gain is concentrated in the spatial and object suites, reflecting larger suite-level variation before convergence.

\begin{table}[!htbp]
\centering
\caption{\libero{} per-suite success rates for the reported settings. Values are success rates in percent over 500 episodes per suite.}
\label{tab:libero-suite-breakdown}
\scriptsize
\setlength{\tabcolsep}{2.5pt}
\begin{tabular}{@{}lclccccc@{}}
\toprule
Setting & Seed & Method & Spatial & Object & Goal & \libero{}-10 & Avg. \\
\midrule
\pifive{} / general & 7 & Baseline & 93.2 & 96.6 & 92.4 & 87.2 & 92.35 \\
\pifive{} / general & 7 & \method{} & \textbf{94.2} & \textbf{98.6} & \textbf{93.0} & \textbf{87.6} & \textbf{93.35} \\
\midrule
\pifive{} / adapted & 1000 & Baseline & 96.4 & 98.0 & 95.6 & 93.0 & 95.75 \\
\pifive{} / adapted & 1000 & \method{} & \textbf{97.8} & \textbf{98.8} & \textbf{97.8} & \textbf{94.0} & \textbf{97.10} \\
\pifive{} / adapted & 7 & Baseline & 96.4 & \textbf{99.0} & 95.8 & \textbf{94.2} & 96.35 \\
\pifive{} / adapted & 7 & \method{} & \textbf{97.0} & 98.6 & \textbf{98.4} & 92.8 & \textbf{96.70} \\
\pifive{} / adapted & 42 & Baseline & \textbf{97.6} & 98.2 & 95.6 & 93.6 & 96.25 \\
\pifive{} / adapted & 42 & \method{} & 97.4 & \textbf{99.0} & \textbf{96.2} & \textbf{94.0} & \textbf{96.65} \\
\pifive{} / adapted & Mean & Baseline & 96.80 & 98.40 & 95.67 & 93.60 & 96.12 \\
\pifive{} / adapted & Mean & \method{} & \textbf{97.40} & \textbf{98.80} & \textbf{97.47} & 93.60 & \textbf{96.82} \\
\midrule
\oft{} / general & 7 & Baseline & 19.8 & 66.0 & \textbf{67.0} & \textbf{17.2} & 42.50 \\
\oft{} / general & 7 & \method{} & \textbf{35.8} & \textbf{89.4} & 58.8 & 11.8 & \textbf{48.95} \\
\bottomrule
\end{tabular}
\end{table}

\subsection{Confidence intervals and paired comparisons}
\label{app:confidence-intervals}

Table~\ref{tab:key-ci} reports Wilson 95\% confidence intervals for the binomial success metrics in the main paper.
For the \calvin{} and \libero{} adapted settings we add a three-seed pooled interval alongside the primary single-seed run, with the per-seed numbers in Tables~\ref{tab:calvin-paired-seeds} and~\ref{tab:libero-suite-breakdown}.
\libero{} intervals pool the four task suites of each comparison.
For SO-101, the all-colors rows pool all 128 physical trials per policy (including the held-out green condition), while the ID-colors row pools only the yellow, orange, and purple trials.

\begin{table}[!htbp]
\centering
\caption{Wilson 95\% confidence intervals for success metrics.}
\label{tab:key-ci}
\scriptsize
\setlength{\tabcolsep}{3pt}
\begin{tabular}{@{}llccc@{}}
\toprule
Evaluation setting & Method & Metric & Value (\%) & 95\% CI (\%) \\
\midrule
\calvin{} / \pifive{} / general & Baseline & SR (5) & 0.4 & [0.2, 1.0] \\
\calvin{} / \pifive{} / general & \method{} & SR (5) & \textbf{1.3} & [0.8, 2.2] \\
\calvin{} / \pifive{} / adapted & \source{} & SR (5) & 62.3 & [59.3, 65.3] \\
\calvin{} / \pifive{} / adapted & Baseline & SR (5) & 61.4 & [58.3, 64.4] \\
\calvin{} / \pifive{} / adapted & \method{} & SR (5) & \textbf{72.1} & [69.2, 74.8] \\
\calvin{} / \pifive{} / adapted, 3-seed pooled & Baseline & SR (5) & 62.1 & [60.4, 63.9] \\
\calvin{} / \pifive{} / adapted, 3-seed pooled & \method{} & SR (5) & \textbf{70.9} & [69.2, 72.5] \\
\midrule
\libero{} / \pifive{} / general & Baseline & SR & 92.35 & [91.10, 93.44] \\
\libero{} / \pifive{} / general & \method{} & SR & \textbf{93.35} & [92.17, 94.36] \\
\libero{} / \pifive{} / adapted, seed 1000 & Baseline & SR & 95.75 & [94.77, 96.55] \\
\libero{} / \pifive{} / adapted, seed 1000 & \method{} & SR & \textbf{97.10} & [96.27, 97.75] \\
\libero{} / \pifive{} / adapted, 3-seed pooled & Baseline & SR & 96.12 & [95.60, 96.58] \\
\libero{} / \pifive{} / adapted, 3-seed pooled & \method{} & SR & \textbf{96.82} & [96.34, 97.23] \\
\libero{} / \oft{} / general & Baseline & SR & 42.50 & [40.35, 44.68] \\
\libero{} / \oft{} / general & \method{} & SR & \textbf{48.95} & [46.76, 51.14] \\
\libero{} / \oft{} / adapted & Baseline & SR & 95.35 & [94.34, 96.19] \\
\libero{} / \oft{} / adapted & \method{} & SR & \textbf{96.15} & [95.21, 96.91] \\
\midrule
SO-101 / \pifive{} / all colors & Baseline & contact & 93.8 & [88.2, 96.8] \\
SO-101 / \pifive{} / all colors & \method{} & contact & \textbf{96.9} & [92.2, 98.8] \\
SO-101 / \pifive{} / all colors & Baseline & lift & 81.2 & [73.6, 87.1] \\
SO-101 / \pifive{} / all colors & \method{} & lift & \textbf{82.8} & [75.3, 88.4] \\
SO-101 / \pifive{} / all colors & Baseline & task success & 72.7 & [64.4, 79.6] \\
SO-101 / \pifive{} / all colors & \method{} & task success & \textbf{78.1} & [70.2, 84.4] \\
SO-101 / \pifive{} / ID colors & Baseline & task success & 81.2 & [72.3, 87.8] \\
SO-101 / \pifive{} / ID colors & \method{} & task success & \textbf{83.3} & [74.6, 89.5] \\
SO-101 / \pifive{} / held-out green & Baseline & task success & 46.9 & [30.9, 63.6] \\
SO-101 / \pifive{} / held-out green & \method{} & task success & \textbf{62.5} & [45.3, 77.1] \\
\bottomrule
\end{tabular}
\end{table}

\begin{samepage}
\paragraph{Paired comparisons.}
\label{app:paired-comparisons}
At the evaluated checkpoints, 95\% confidence intervals are obtained by paired percentile bootstrap within \libero{} tasks and over \calvin{} chains.
Binary outcomes are compared using exact two-sided McNemar tests, with Holm correction across the four \libero{} comparisons.
After Holm correction, \pifive{}-adapted ($p=0.0476$) and \oft{}-general \mbox{($p<10^{-7}$)} remain significant.
\par
\end{samepage}
\begin{table}[!htbp]
\centering
\caption{Paired \libero{} success-rate differences (\method{} minus task-loss-only fine-tuning), in percentage points. Each configuration contains 2,000 outcomes paired by task and initial state.}
\label{tab:paired-libero}
\begin{tabular}{lrr}
\toprule
Model / initialization & Success-rate difference & 95\% CI \\
\midrule
\pifive{} / general checkpoint & +1.00 & [-0.50, 2.50] \\
\pifive{} / adapted checkpoint & +1.35 & [0.30, 2.40] \\
\oft{} / general checkpoint & +6.45 & [4.50, 8.35] \\
\oft{} / adapted checkpoint & +0.80 & [-0.10, 1.70] \\
\bottomrule
\end{tabular}
\end{table}
\begin{table}[!htbp]
\centering
\caption{Paired \calvin{} comparisons for \pifive{}. Differences are \method{} minus task-loss-only fine-tuning. Confidence intervals describe average-sequence-length differences; the unadjusted $p$ values refer to SR(5).}
\label{tab:paired-calvin}
\begin{tabular}{lrrr}
\toprule
Initialization & Difference in average sequence length & 95\% CI & SR(5) $p$-value \\
\midrule
General checkpoint & +0.216 & [0.135, 0.297] & 0.0490 \\
Adapted checkpoint & +0.279 & [0.171, 0.388] & $1.13\times10^{-10}$ \\
\bottomrule
\end{tabular}
\end{table}
\subsection{CALVIN low-learning-rate calibration point}
\label{app:calvin-lowlr}

Table~\ref{tab:calvin-lowlr} reports a low-learning-rate \calvin{} calibration point.
It shares the adapted starting checkpoint, CALVIN ABC training data, and CALVIN chain-evaluation procedure of the main \calvin{} row, but trains for 4{,}000 steps at a constant learning rate of $6.25{\times}10^{-6}$.

\begin{table}[!htbp]
\centering
\caption{\calvin{} low-learning-rate constant-schedule calibration point.}
\label{tab:calvin-lowlr}
\small
\begin{tabular}{@{}lccccc@{}}
\toprule
Starting checkpoint & Method & Step & Avg.\ seq.\ len. & SR (5) (\%) & Trials \\
\midrule
CALVIN adapted &Baseline & 4,000 & 3.992 & 63.9 & 1000 \\
CALVIN adapted &\method{} & 4,000 & \textbf{4.206} & \textbf{71.5} & 1000 \\
\bottomrule
\end{tabular}
\end{table}

\subsection{Published \libero{} results and continued fine-tuning}
\label{app:published-model-context}

Table~\ref{tab:external-context} places our benchmark settings alongside published \libero{} results, with sources and evaluation settings noted for each reference point.
The final block reports evaluation of the released LeRobot \pifive{} \libero{} adapted checkpoint \texttt{lerobot/pi05\_libero\_finetuned}, together with Baseline and \method{} results after 300 steps of continued fine-tuning from this near-saturated checkpoint under the same training and evaluation conditions.

\begin{table}[!htbp]
\centering
\caption{\libero{} reference points from published VLA results and continued fine-tuning of an adapted policy. Values are success rates in percent over the four \libero{} suites when suite-level results are available.}
\label{tab:external-context}
\scriptsize
\setlength{\tabcolsep}{1.8pt}
\begin{tabularx}{\linewidth}{@{}YcccccY@{}}
\toprule
Reference point & Spatial & Object & Goal & \libero{}-10 & Avg. & Source / note \\
\midrule
Diffusion Policy (scratch) & 78.3 & 92.5 & 68.3 & 50.5 & 72.4 & \citep{chiDiffusionPolicy2023,kimOpenVLAOpenSourceVisionLanguageAction2024} \\
Octo (fine-tuned) & 78.9 & 85.7 & 84.6 & 51.1 & 75.1 & \citep{octo2024,kimOpenVLAOpenSourceVisionLanguageAction2024} \\
OpenVLA (fine-tuned) & 84.7 & 88.4 & 79.2 & 53.7 & 76.5 & \citep{kimOpenVLAOpenSourceVisionLanguageAction2024} \\
$\pi_0$ + FAST (fine-tuned) & 96.4 & 96.8 & 88.6 & 60.2 & 85.5 & \citep{pertschFAST2025,blackPi0VisionLanguageActionFlow2024,kimFineTuningVisionLanguageActionModels2025} \\
$\pi_0$ (fine-tuned) & 96.8 & 98.8 & 95.8 & 85.2 & 94.2 & \citep{blackPi0VisionLanguageActionFlow2024,kimFineTuningVisionLanguageActionModels2025} \\
\oft{} with additional inputs & 97.6 & 98.4 & 97.9 & 94.5 & 97.1 & \citep{kimFineTuningVisionLanguageActionModels2025} \\
GR00T N1.5 & 98.2 & 99.4 & 97.2 & 87.8 & 95.7 & \citep{bjorckGR00TN1Open2025,kimContrastiveRepresentationRegularization2025} \\
GR00T N1.5 + RS-CL & 98.4 & 98.6 & 98.2 & 90.4 & 96.4 & \citep{kimContrastiveRepresentationRegularization2025,bjorckGR00TN1Open2025} \\
\midrule
\makecell[l]{\pifive{} adapted source\\\texttt{lerobot/pi05\_libero\_finetuned}} & 98.4 & 99.0 & 96.0 & 95.4 & 97.20 & Released checkpoint evaluation \\
\makecell[l]{\pifive{} continued fine-tuning\\Baseline} & 97.6 & 99.0 & 97.6 & 95.2 & 97.35 & Step 300 \\
\makecell[l]{\pifive{} continued fine-tuning\\\method{}} & 97.6 & 99.2 & 97.8 & 96.0 & \textbf{97.65} & \textbf{+0.30 pp / +0.31\% vs Baseline} \\
\bottomrule
\end{tabularx}
\end{table}

\subsection{SO-101 trial breakdown}
\label{app:so101-trial-breakdown}

Table~\ref{tab:so101-color-breakdown} breaks down the 128 physical trials per policy by block color: yellow, orange, and purple are in-distribution colors from the 90-demonstration training set, and green is held out for color generalization.
These rows give the per-color breakdown behind the aggregate results; the in-distribution colors remain high overall, and the held-out green condition shows the largest task-success gain.

\begin{table}[!htbp]
\centering
\caption{SO-101 physical trial breakdown by color. Entries report successes out of 32 trials per color for each policy, with percentages in parentheses.}
\label{tab:so101-color-breakdown}
\scriptsize
\setlength{\tabcolsep}{3pt}
\begin{tabular}{@{}lllccc@{}}
\toprule
Color & Split & Method & Contact & Lift & Task success \\
\midrule
Yellow & ID & Baseline & 30/32 (93.8) & 25/32 (78.1) & 25/32 (78.1) \\
Yellow & ID & \method{} & \textbf{32/32 (100.0)} & \textbf{28/32 (87.5)} & \textbf{26/32 (81.2)} \\
Orange & ID & Baseline & 32/32 (100.0) & 29/32 (90.6) & 26/32 (81.2) \\
Orange & ID & \method{} & 32/32 (100.0) & 29/32 (90.6) & \textbf{28/32 (87.5)} \\
Purple & ID & Baseline & 31/32 (96.9) & \textbf{27/32 (84.4)} & \textbf{27/32 (84.4)} \\
Purple & ID & \method{} & \textbf{32/32 (100.0)} & 26/32 (81.2) & 26/32 (81.2) \\
Green & OOD & Baseline & 27/32 (84.4) & 23/32 (71.9) & 15/32 (46.9) \\
Green & OOD & \method{} & \textbf{28/32 (87.5)} & 23/32 (71.9) & \textbf{20/32 (62.5)} \\
\bottomrule
\end{tabular}
\end{table}

\subsection{Validation of the shared action probe}
\label{app:probe-validation}
\paragraph{Held-out prediction and row-space stability.}
On \oft{}, saved online probes at 5k, 10k, 15k, and 20k training steps explain 49--56\% of held-out action variance without refitting.
Consecutive probe row spaces have mean principal cosine 0.800--0.884, averaged over their seven principal angles.
\paragraph{Action information conditional on nonvisual inputs.}
For the 20k \oft{} checkpoint, we fix the saved probe's row space and evaluate its seven orthonormal coordinates using two-fold cross-fitted ridge readouts conditioned on instruction identity and proprioception.
These coordinates retain 94.0\% of the full visual representation's additional linear action $R^2$, measured relative to a context-only readout.
Their additional $R^2$ is 6.71 times the mean of 32 rank-matched random subspaces and exceeds every control.
\paragraph{Sensitivity comparison.}
On \calvin{} with \pifive{}, we compare the shared probe with
\begin{equation}
G=\mathbb E\!\left[(J\Sigma_R^{1/2})^\top(J\Sigma_R^{1/2})\right],
\end{equation}
where $\Sigma_R$ is the frozen feature covariance.
We consider Jacobians $J$ of the final decoded action, individual denoising-path increments, and teacher-forced velocity predictions.
Across these three definitions, 48.7--49.7\% of the activation-weighted probe-operator energy lies in the corresponding top-seven sensitivity subspace.
The fraction of activation-weighted Jacobian energy captured by the probe row space is 35--39 times the mean for 256 rank-matched random subspaces, exceeding every control in each case and supporting the action-predictive feature-channel approximation.

\subsection{Residual evolution and visual content}
\label{app:residual-diagnostics}

\paragraph{Evolution on fixed states.}
We evaluate residual geometry at initialization, 3,000 steps, and the final checkpoint on a fixed set of 64 states under the \calvin{}-general setting, using checkpoint-specific cross-fitted probes.
Relative to initialization, final-checkpoint residual effective rank and CKA decrease under task-loss-only fine-tuning and increase with \method{} (Table~\ref{tab:residual-evolution}).
\begin{table}[!htbp]
\centering
\caption{Residual participation-ratio effective rank and teacher CKA on fixed \calvin{}-general states.}
\label{tab:residual-evolution}
\begin{tabular}{llrr}
\toprule
Method & Checkpoint & Effective rank & Teacher CKA \\
\midrule
Task-loss only & Initialization & 11.35 & 0.255 \\
Task-loss only & 3,000 steps & 3.80 & 0.153 \\
Task-loss only & Final checkpoint & 4.24 & 0.112 \\
\method{} & Initialization & 11.35 & 0.255 \\
\method{} & 3,000 steps & 58.31 & 0.433 \\
\method{} & Final checkpoint & 76.61 & 0.481 \\
\bottomrule
\end{tabular}
\end{table}

\paragraph{Paired color interventions.}
The intervention fixes the command, robot state, geometry, and optimal-action label while varying block color.
Response energy is the summed squared displacement from the mean over six color permutations on a $4\times4$ visual-token grid.
The training palette uses red, blue, and pink; the held-out palette uses green, yellow, and cyan.
The probe-predictive fractions fall below all 64 isotropic rank-seven random controls, whereas dimension-normalized residual/component ratios exceed every control, localizing the amplified color response non-randomly to the probe-filtered residual.
\begin{table}[!htbp]
\centering
\caption{Color-response diagnostics for training and held-out palettes. The last two rows use the saved \method{} probe. The energy fraction is relative to full-token response energy; the per-dimension ratio is $(E_R/(d-7))/(E_P/7)$, where $E_R$ and $E_P$ are residual and probe-predictive response energies and $d$ is the channel dimension.}
\label{tab:color-response}
\begin{tabularx}{\linewidth}{Xcc}
\toprule
Measure & Training palette & Held-out palette \\
\midrule
Color-response energy ratio (\method{} / task-loss-only) & $1.889\times$ & $2.067\times$ \\
Probe-predictive energy fraction & $0.2208\%$ & $0.1997\%$ \\
Per-dimension energy ratio (residual / probe-predictive component) & $1.550\times$ & $1.714\times$ \\
\bottomrule
\end{tabularx}
\end{table}

\paragraph{Dimension-matched semantic readout.}
A ridge readout uses four-fold cross-validation split by scene condition, fitting a residual subspace of at most seven dimensions within each training fold and comparing it with seven orthonormal coordinates of the saved probe row space.
The five predeclared targets are the locations of the red, blue, and pink blocks and the occupants of the left and right slider slots.
Macro balanced accuracy is 77.09\% in the \method{} residual and 51.16\% in its probe-predictive component.
The difference is +25.93 pp (95\% CI [23.86, 27.79]), and the residual readout exceeds all 64 random residual controls.

\paragraph{Transfer across palettes.}
A separate ridge readout predicts drawer, LED, lightbulb, and slider states from globally mean-pooled features, training on the training palette and evaluating on the held-out palette with disjoint scene conditions.
Macro balanced accuracy is 89.71\% after task-loss-only fine-tuning and 100.00\% with \method{}, a +10.29 pp difference (95\% CI [9.77, 10.76]).
For \method{}, residual and probe-predictive component accuracies are 100.00\% and 71.95\%, respectively.

\subsection{Additional filtering ablations}
\label{app:component-details}

We compare full-token, random-subspace, and probe-filtered variants using rank regularization without teacher alignment.

\begin{table}[!htbp]
\centering
\caption{\textbf{Filtering ablations on \calvin{}.} Values are SR(5) in percent.}
\label{tab:calvin-filter-controls}
\begin{tabular}{lr}
\toprule
Variant & SR(5) (\%) \\
\midrule
Full-token rank only & 65.1 \\
Random-subspace rank only & 60.5 \\
Residual rank only & \textbf{70.5} \\
\bottomrule
\end{tabular}
\end{table}

The higher success rate with learned filtering supports its contribution to the gains from residual rank regularization.

\end{document}